\documentclass[letterpaper, 10 pt, conference]{ieeeconf}  

\IEEEoverridecommandlockouts                              

\usepackage{amsmath} 
\usepackage{amssymb}  
\usepackage{dsfont}
\usepackage{graphicx}
\usepackage{subfigure}
\usepackage{psfrag,graphicx,epsfig}
\usepackage{epstopdf}
\usepackage{xspace}
\usepackage{subfig}
\usepackage{float}
\usepackage{placeins}
\usepackage{multirow}
\usepackage{pgf,tikz}
\usepackage{nowidow}
\usepackage{lineno}
\usepackage{xcolor}

\usepackage{siunitx}
\usepackage{color}
\usepackage{flushend}
\usepackage{lineno}
\usepackage{algorithmic}

\newtheorem{thm}{Theorem}[section]

\newtheorem{assumption}{Assumption}

\allowdisplaybreaks

\title{\LARGE \bf
Design of Adaptive Compliance Controllers\\ for Safe Robotic Assembly
}

\author{Devesh K. Jha, Diego Romeres, Siddarth Jain, William Yerazunis and Daniel Nikovski$^{\dagger}$,%
\thanks{$^{\dagger}$All authors are with Mitsubishi Electric Research Laboratories (MERL), Cambridge, MA, USA 02139 {\tt\small \{jha,romeres,sjain,yerazunis,nikovski\}@merl.com}}
}%

\begin{document}

\maketitle


\begin{abstract}
Insertion operations are a critical element of most robotic assembly operation, and peg-in-hole (PiH) insertion is one of the most widely studied tasks in the industrial and academic manipulation communities. PiH insertion is in fact an entire class of problems, where the complexity of the problem can depend on the type of misalignment and contact formation during an insertion attempt. In this paper, we present the design and analysis of adaptive compliance controllers which can be used in insertion-type assembly tasks, including learning-based compliance controllers which can be used for insertion problems in the presence of uncertainty in the goal location during robotic assembly. We first present the design of compliance controllers which can ensure safe operation of the robot by limiting experienced contact forces during contact formation. Consequently, we present analysis of the force signature obtained during the contact formation to learn the corrective action needed to perform insertion. Finally, we use the proposed compliance controllers and learned models to design a policy that can successfully perform insertion in novel test conditions with almost perfect success rate. We validate the proposed approach on a physical robotic test-bed using a 6-DoF manipulator arm.  
\end{abstract}
\section{Introduction}\label{sec:introduction}
Over the last several decades, robots have become very precise in performing repetitive pick-and-place operations. However, complications arise when the positions of the parts involved in assembly vary between repetitions of the operation. A classical example of such a task is PiH insertion, which has been studied extensively in assembly for a long time, due to its relevance to manufacturing~\cite{xu2019compare}. This task is a major component of a lot of assembly operations. Even though this task has been studied in robotics and automation research for a long time, this problem remains open in multiple aspects. Presence of pose uncertainty for the parts being assembled leads to complex contact configurations between the parts.  Consequently, manipulation for successful assembly requires design of force-feedback controllers that can interpret contact forces and correct the contact configuration, so that the parts could be assembled. Since these contact configurations depend on the physical, as well as the geometrical features of the objects being assembled, they are notoriously difficult to model precisely. As a result, learning-based approaches have been very popular for designing the required corrective controllers. However, learning-based approaches also present challenges when it comes to designing efficient controllers which can reliably perform assembly in the presence of sustained contact interactions. 

\begin{figure}
    \centering
    \includegraphics[width=0.35\textwidth]{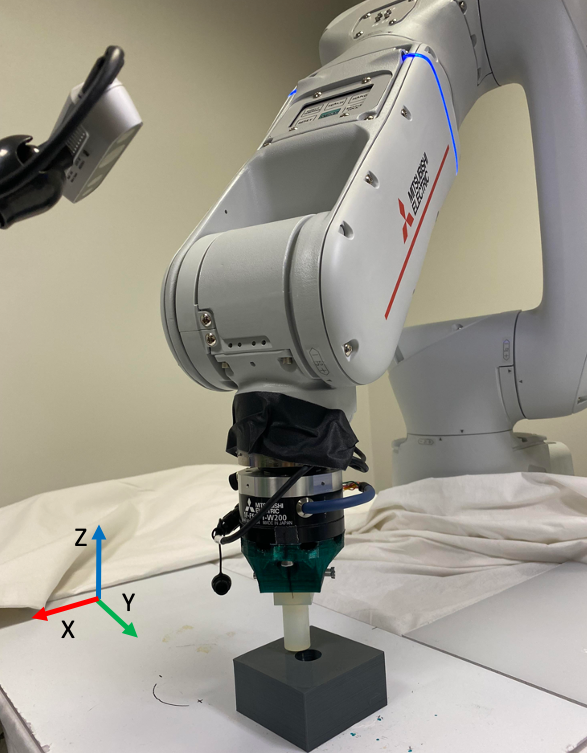} %
    \caption{Experimental setup with a Mitsubishi Electric Factory Automation (MELFA) RV-$5$AS-D Assista $6$  DoF manipulator arm in a possible contact configuration with the hole environment. The diameter of the peg is approximately $20$ mm and the tolerance is approximately $1.0$ mm. The figure also shows an Intel Realsense D435 camera which is used in the experiments for the detection of the hole.}
    \label{fig:exp_setup}
\end{figure}

The design of learning-based controllers requires an initial exploration phase where the robot has to explore different possible contact configurations so that a generalizable corrective policy could be learnt. We believe that one of the key components of the design of such controllers is the design of compliant controllers which can ensure safe interaction of the robot with the assembly components during the exploration phase. Whereas this is a key requirement for learning adaptive assembly controllers, design of such safe controllers remains mostly unexplored. With this motivation, we present the design of a class of accommodation controllers which guarantee that the contact forces will remain within safe bounds. This accommodation controller is used to collect contact force data to learn a relationship between misalignment and the expected contact forces. To interpret the data efficiently, we present analysis of the contact forces for the purpose of selecting features which can be used to learn a predictive model between contact forces and the amount of expected misalignment. Finally, this predictive model is used to design a corrective policy that allows assembly using the predictive model.

\textbf{Contributions.} This paper has the following key contributions:
\begin{enumerate}
    \item We present the design and analysis of two different controllers for safe interaction between the robot and its environment in the presence of sustained contacts. 
    \item We present feature analysis for the design of an efficient force-feedback controller for interpretation of different complex contact configurations. 
    \item We present the design and verification of a learning-based controller that makes use of the proposed safe accommodation controller and the proposed feature analysis for insertion-type assembly using a $6$-DoF manipulator system.
\end{enumerate}
\section{Related Work}
\label{sec:RelatedWork}
Automatic assembly is one of the most common robot applications, and what differentiates it from many other robotic applications is the need to carefully consider the effect of contact between assembled parts. Pure position control is usually inadequate, because if the robot uses such a controller to follow a reference trajectory exactly, even small misalignments between where the assembled parts are and where they were expected to be would result in very large forces, possibly damaging the robot and/or the parts. A much more suitable type of control for this application is force control that adjusts the motion of the robot in response to experienced contact forces. Such methods for robotic assembly are commonly known as{\em adaptive assembly strategies} (AAS) (\cite{Nevins1977ResearchAutomation,Gullapalli1994AcquiringLearning,Abu-Dakka2015AdaptationProfiles,Levine2016End-to-endPolicies,dong2021icra,Kulkarni2021LearningLearning, DBLP:journals/corr/abs-2007-11646}). A lot of the research in this area has focused on the PiH insertion problem, as a prototypical operation for various assembly tasks.

The main challenge in an AAS is how to interpret the measured force/torque (F/T) signals in order to direct the motion of the robot so as to accomplish the insertion. As early as the 1970s, it was successfully demonstrated that high-accuracy PiH insertion was possible by direct interpretation of F/T signals by a robot program \cite{Inoue1974ForceTasks}. However, the development of such robot programs is very complex, laborious, expensive, and case-dependent, so this approach turned out to be impractical for wide industrial use. A more universally-applicable approach is to follow a suitable position trajectory that would accomplish the task in the absence of collisions, and adjust the robot's motion in response to contact forces, thus forming a force feedback controller \cite{Nevins1977ResearchAutomation}. Many such controllers use a mapping from the F/T readings measured at the wrist of the robot (or the platform that the hole is mounted on), onto a correction to the trajectory. In some rare instances, this mapping can be computed analytically -- for example, when the peg and hole are circular, have no angular misalignment, overlap at least to some extent, and the point around which the moments of the F/T sensor are computed lies on the axis of the peg (which is also the direction of insertion, \cite{Gottschlich1989AMating}). However, this kind of solution requires careful placement and alignment of the F/T sensor, and is not general enough for regular use.


As this approach to designing force controllers reduces to finding a suitable mapping between F/T measurements and corrections to a nominal trajectory, a much more general method for obtaining this mapping, and thus a working controller, is to use machine learning methods for estimating the mapping from data. One early such method for {\em programmed compliance} proposed in \cite{Peshkin1990ProgrammedAssembly} used linear least squares to estimate a linear mapping between F/T measurements and corrections to either position or velocity, effectively learning the admittance and accommodation matrices used in linear compliance controllers. The training examples needed for learning were constructed based on general considerations about what the corrections should be for prototypical situations, and what contact forces might be measured in them. However, it was later demonstrated that for the contact configurations usually experienced in PiH insertion tasks, the mapping between forces and corrections was not linear \cite{Asada1990TeachingCompliance,Asada1993RepresentationNets}, and it was suggested to use neural networks to represent a non-linear mapping between the two. This advance significantly expanded the type of mappings that could be learned, but still left open the question of how a suitable data set of training examples could be compiled, as doing this manually is excessively difficult for all but the simplest geometries. 
A much more appealing solution is to measure contact forces directly on a real robot by putting the peg and hole in various contact situations. Gullapalli et al. (\cite{Gullapalli1994AcquiringLearning} proposed a reinforcement learning (RL) solution based on trial and error, which learned to associate the contact forces with a correction that was advantageous in bringing the peg closer to its desired end position, while also minimizing contact forces. The desired outcome was encoded in the reward function of the RL problem formulation. Although this approach achieved remarkable results, learning to insert a peg in a hole with clearances significantly lower than the accuracy (repeatability) of the robot used, this method still needed accurate knowledge of where the goal position was, in order to use it in the reward function. This precludes its direct use in the version of the problem we consider, where the uncertainty is precisely in the position of the hole, and thus in the correct end position that the peg should reach.    

Following this seminal application of RL to PiH insertion, a number of later works explored the use of machine learning models for the design of adaptive force controllers. The application of deep RL for learning end-to-end visuomotor policies was demonstrated in \cite{Levine2016End-to-endPolicies}. In addition to F/T sensors, tactile sensors have been employed, too (\cite{Dong2019Tactile-BasedBox-Packing, dong2021icra}). As the instantaneous F/T readings might not be sufficient to disambiguate the contact configuration, the use of recurrent neural nets has been proposed in \cite{Inoue2017DeepTasks,Kulkarni2021LearningLearning}. However, as is well known, RL often suffers from unfavorable sample complexity, making it less suitable for use on real mechanical systems. In contrast, we explore below a supervised learning approach to learning mappings between forces and corrections, thus significantly reducing the number of training samples needed. 

The work proposed in this paper is closest to our previous work in~\cite{jha2021imitation}. However, compared to the work in~\cite{jha2021imitation},  we present design of an additional nonlinear accommodation controller, we present a proof for convergence of the controller, as well as feature analysis for threshold detection. Furthermore, we show an improvement in the final insertion system which uses a DL-based hole detection method along with a faster controller for insertion.

\section{Problem Statement}\label{sec:problem_statement}
\begin{figure}
    \centering
    \includegraphics[width=0.40\textwidth]{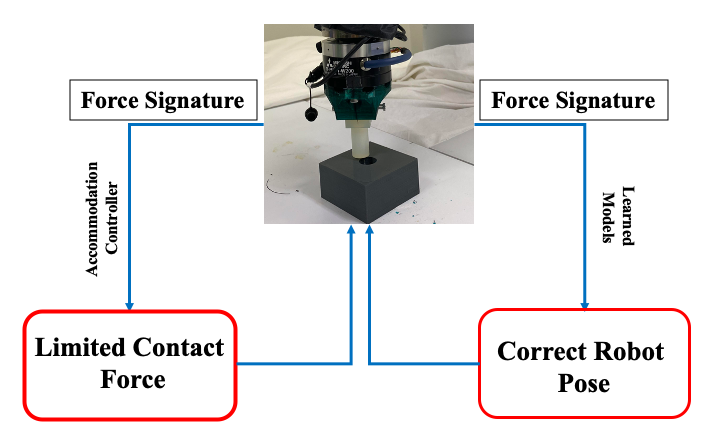} %
    \caption{The control system design that we study in this paper. The important point to note is that the force feedback is used to design the accommodation controller as well as to correct the object pose for successful insertion.}
    \label{fig:pih_analytical}
\end{figure}

In this section, we present the problem that we are trying to solve in this paper. Loosely speaking, the objective is to control the contact state and the problem state (i.e., the pose of the peg for our case) during an insertion attempt. 
The schematic in Figure~\ref{fig:pih_analytical} shows the twofold objective of using force feedback in controller design for assembly. The force feedback is used to design a lower-level controller to limit interaction forces in the event of contact formation during an insertion attempt. 
As could be seen in the figure, any insertion attempt leads to a contact formation and the goal is to use the corresponding force signature to correct for the underlying misalignment. However, we would like that the interaction forces obtained during any arbitrary contact formation be bounded, irrespective of the reference trajectory provided to the robot. Furthermore, we would like to learn models using quasi-steady behavior of the system for ease of learning and prediction. 

In all all of these cases, we assume that the misalignment is only in a plane, and that there is no angular misalignment between the peg and the hole. This corresponds to the often encountered case in practice when the hole base slides across a working surface in a factory, for example a workbench. In order to train a model, we try to solve the following problems in this paper, which are then used together to design a force-feedback controller for performing peg-in-hole assembly in the presence of significant positional inaccuracy. 
\begin{enumerate}
    \item Suppose that we have a reference trajectory for insertion, which is denoted as $X_\text{insertion}=\{x_r[0],x_r[1],\dots,x_r[k],\dots,x_r[N]\}$. Suppose that due to contact formation, the robot experiences a sequence of contact forces denoted as $F_\text{insertion}=\{f[0],f[1],\dots,f[k],\dots,f[N]\}$. The force control task is to design a force feedback controller that  modifies $X_\text{insertion}$ using a force feedback law so that $\exists $ $\mathcal{K} \in \mathbb{Z}$ such that $\forall$ $k\geq \mathcal{K}$, $||f[k]-f[k-1]||_2\leq \epsilon$, where $\epsilon$ is arbitrarily small.
    \item The second task is to then analyze and use the force signature data obtained from the force controller to design a force feedback controller to correct the misalignment between the peg and hole position. 
\end{enumerate}
In summary, the goal is to use force feedback to design both the lower level accommodation controller, as well as the corrective policy that can allow the robot to correct any contact formation for successful assembly (see also Figure~\ref{fig:pih_analytical}). 

\section{Controller Design}\label{sec:controller_design}
In this section, we present the design and analysis of the compliance controllers that we use to ensure safe interaction during insertion. We believe that this is a critical step to ensure safety of the learning process. Even though there has recently been a lot of work in robot learning approaches for performing manipulation, ensuring the safety of the contact-rich interactions during these tasks has largely been overlooked. However, this is a very critical requirement for adoption of these learning-based approaches for use in assembly operations, and many other related operations. Based on this motivation, we present the design and analysis of two different kinds of controllers using force feedback with different convergence behaviors. 

\begin{figure}
    \centering
    \includegraphics[width=0.40\textwidth]{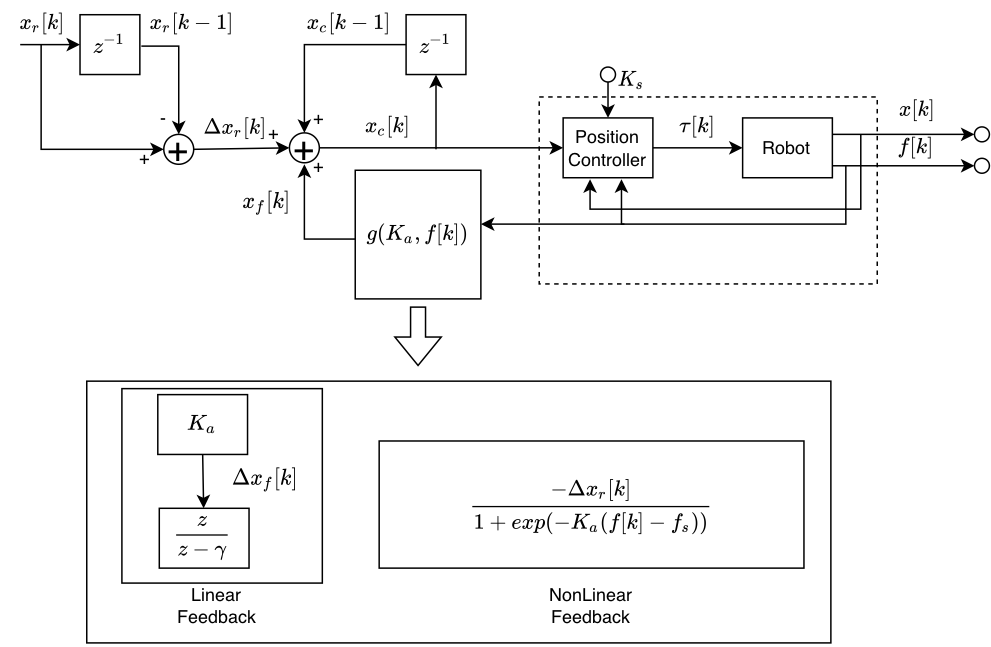} %
    \caption{Block diagrams for the two controllers described in the paper. $K_a$: Accommodation Matrix, $K_s$: Stiffness Gains of the low-level stiffness controller.}
    \label{fig:gen_acc_block_diagram}
\end{figure}

In both of these controllers, we use the force measured by a force-torque sensor mounted at the wrist of the robot (see Figure~\ref{fig:exp_setup}) to adapt a reference trajectory to regulate interaction forces experienced by the robot with their environment. The idea is to use force feedback to modify the reference trajectory so as to limit the contact forces to allowable bounds. For clarity of presentation, we present block diagrams for both controllers in Figure~\ref{fig:gen_acc_block_diagram}. 



\subsection{Linear Accommodation Controller Design}\label{subsec:linear_accommodation_controller}
The operation of the proposed linear accommodation controller is presented in Figure~\ref{fig:gen_acc_block_diagram}. As could be seen in the block diagram in Figure~\ref{fig:gen_acc_block_diagram}, the accommodation controller modifies the reference trajectory using force feedback. In particular, the accommodation controller uses the following feedback law to modify the reference trajectory of the robot. Let us denote the discrete-time reference trajectory by $x_r[k]$, the trajectory commanded to the low-level position controller by $x_c[k]$, the experienced forces by $f[k]$, the measured position as $x[k]$, at any instant $k$. Note that the $k$ here denotes the control time index and not the actual time in seconds. In this design, we employ a low-level compliant position controller that makes the robot behave like a spring-damper system with desired stiffness and damping coefficients. Most robot vendors provide such a stock controller with the robot, or if not, one can be implemented relatively easily (\cite{Lynch2017ModernRobotics}). Let us denote the stiffness constant of the compliant position controller by $K_s$ and the accommodation matrix for the force feedback by $K_a$. For simplicity, we consider a diagonal matrix $K_a$. With this assumption, we present the force-feedback law for updating the commanded position along each individual axis next. The commanded trajectory sent to the robot is computed using the following update rule (also see Figure~\ref{fig:gen_acc_block_diagram}):
\begin{equation}\label{eqn:linear_accommodation_force_feedback}
      x_c[k]=x_c[k-1]+\Delta x_r[k]+\sum_{i=0}^{k-1}\gamma^{k-i}K_a f[i]
\end{equation}
where $\gamma \in (0,1)$ is a discounting parameter for computing the integral error, and $\Delta x_r[k]=x_r[k]-x_r[k-1]$ are desired position increments computed from the reference trajectory. An actual force trajectory obtained for a reference trajectory that advances with constant velocity along the $z$-axis of the robot under the operation of the linear accommodation controller is shown in Figure~\ref{fig:force_convergence}a. Note that even though the reference trajectory $x_c[k]$ keeps advancing, the experienced force stabilizes; this behavior is in contrast to that of the stock compliant controller, where contact forces grow proportionally to the advance of the reference position, and can easily become dangerously large for the robot or manipulated parts. (It is generally not feasible to limit these forces by making the stiffness $K_s$ of the stock compliant controller very low, because the robot does not know exactly where an obstacle will be encountered. In contrast, the proposed accommodation controller guarantees bounded forces, even if the reference trajectory advances to infinity, as long as this happens at a constant velocity. The latter condition can be guaranteed easily by sampling any desired geometric reference trajectory accordingly.)

\subsection{Non-linear Accommodation Controller Design}\label{subsec:nonlinear_accommodation_controller}

\begin{figure}[htp]
  \centering
  \subfigure[Force signature obtained by the Linear Accommodation Controller]{\includegraphics[scale=0.25]{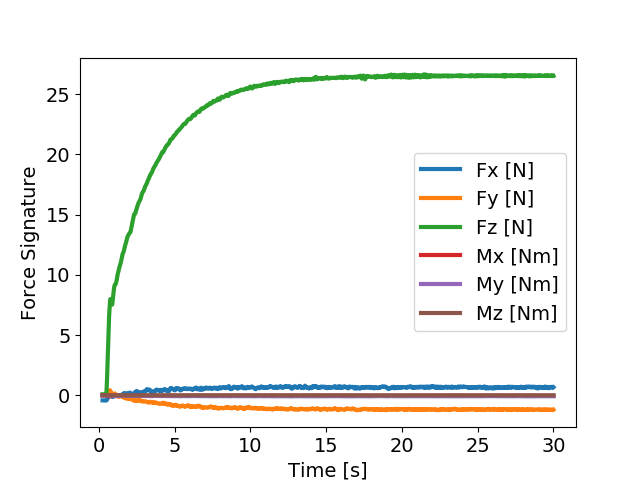}}\quad
  \subfigure[Force signature obtained by the Nonlinear Accommodation Controller]{\includegraphics[scale=0.25]{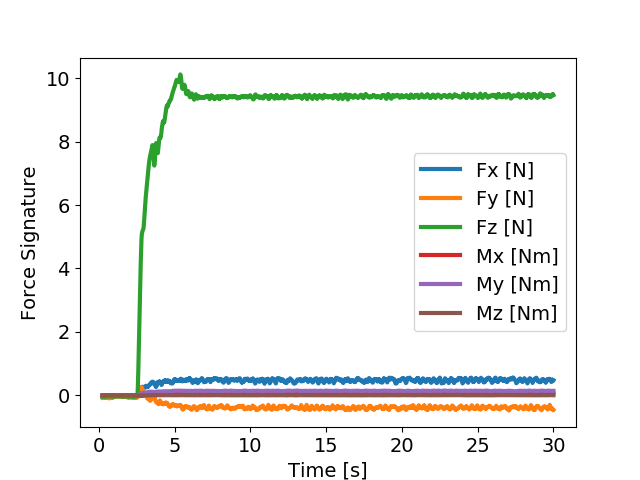}}
  \caption{Force trajectories obtained by the same reference trajectory, but using the two different accommodation controllers from Section~\ref{sec:controller_design}. As we can observe from the plots above, the non-linear force controller can achieve much faster convergence, and the forces converge to lower values in the presence of the same reference trajectory.}
  \label{fig:force_convergence}
\end{figure}
Next, we present a nonlinear feedback law to design an accommodation controller. The corresponding block diagram for this controller is shown in Figure~\ref{fig:gen_acc_block_diagram}. Using the nomenclature from Section~\ref{subsec:linear_accommodation_controller}, the non-linear force feedback law is given by the following equation (also see Figure~\ref{fig:gen_acc_block_diagram}):
\begin{equation}\label{eqn:nonlinear_accommodation_force_feedback}
    x_c[k]=x_c[k-1]+\Delta x_r[k]-\alpha[k]\Delta x_r[k],
\end{equation}
where $\alpha[k]=\frac{1}{(1+\exp(-K_a(f[k]-f_\text{sat})))}$, where $f_\text{sat}$ is specified by the user and approximately defines the force around which the controller would converge. Note that the control law in~\ref{eqn:nonlinear_accommodation_force_feedback} does not use an integration block. Rather, the idea here is to use the the force feedback to cancel any increment of the commanded trajectory. The proposed feedback law ensures that the feedback does not interfere with the movement of the robot in free space (as the force feedback is close to zero in free space). However, since $\alpha[k]$ quickly converges to $1$ as forces go beyond $f_\text{sat}$, this would lead to convergence of the commanded trajectory and hence of the contact forces. The convergence behavior could also be seen in the plot for the nonlinear accommodation controller in Figure~\ref{fig:force_convergence}b.

\subsection{Convergence Analysis}\label{subsec:convergence_analysis}

Next, we state a Theorem which proves that under the assumption of constant velocity of trajectories, the interaction forces will converge for the controller presented in Section~\ref{subsec:linear_accommodation_controller} and~\ref{subsec:nonlinear_accommodation_controller}. To prove convergence of forces, we will need an assumption that relates the robot position and the commanded position with contact forces.
\begin{assumption}\label{assumption1}
The robot is equipped with a stiffness controller with stiffness constant $K_s$ such that the forces observed during an interaction is given by $f_\text{obs}=K_s(x-x_c)+\delta$, where $x$, $x_c$ and $\delta$ are the robot actual state, robot commanded state and observation noise respectively.
\end{assumption}
We make another assumption regarding the velocity of the reference trajectory of the robot.
\begin{assumption}\label{assumption2}
The reference trajectory of the robot has a constant velocity, i.e., $\Delta x_r[k]=\Delta x_r$ $\forall k \in \mathbb{Z}$.
\end{assumption}
With these two assumptions, we can now state the following theorem.
\begin{thm}
Suppose that a reference trajectory with constant velocity is modified with the force feedback specified in Equations~\eqref{eqn:linear_accommodation_force_feedback} and~\eqref{eqn:nonlinear_accommodation_force_feedback}. Suppose that the robot makes contact with a rigid environment at time instant $\mathcal{J}\in \mathbb{Z}$. Then, we have that the following is true: $\exists$ a $\mathcal{K} > \mathcal{J}\in \mathbb{Z}$ such that $\forall$ $k\geq \mathcal{K}$, $||f[k]-f[k-1]||_2\leq \epsilon$, where $\epsilon$ can be made arbitrarily small. 

\begin{proof}
 Since the robot moves with constant velocity in free space, there is no force experienced by the force sensor (except for the measurement noise). Thus, we ignore the part of the trajectory before contact formation. 
 
 Upon contact formation with the external environment, using Assumption~\ref{assumption1} (we ignore the noise term) and Equation~\eqref{eqn:linear_accommodation_force_feedback}, we get the following:
\begin{align}
    f[k]-f[k-1] &= K_s(x-x_c[k])-K_s(x-x_c[k-1]) \nonumber \\
     &= K_s(x_c[k-1]-x_c[k])\nonumber \\
     &=K_s(-\Delta x_r -\sum_{i=0}^{k-1}\gamma^{k-i}K_a f[i]) 
\end{align}

For simplicity of notation, let us denote the summation term by $e[k]$. Thus, the above equation can be simplified as follows:
\begin{align}
f[k]-f[k-1] &=-K_s(\Delta x_r -e[k])
\end{align}
Using the above equation, we can write that $||f[k]-f[k-1]||_2 =K_s||(\Delta x_r -e[k])||_2$. Note that $e[k]$ is a discounted infinite sum of the sequence of observed forces times a gain term. To show convergence, we make an assumption that we can find at least one $\gamma \in (0,1)$ and accommodation term $K_a$,  such that $||(\Delta x_r -e[k])||_2 \leq \eta$, $\forall k >\mathcal{K}$, where $\eta$ is arbitrarily small. Using this assumption, then we have that $||f[k]-f[k-1]||_2 \leq K_s\eta \leq \epsilon$. 

Convergence of the nonlinear controller given by Equation~\eqref{eqn:nonlinear_accommodation_force_feedback} is straightforward. It can be shown by the convergence properties of $\alpha[k]$. We show this in the following text. Equation~\eqref{eqn:nonlinear_accommodation_force_feedback} can be re-arranged as following:
\begin{equation}\label{eqn:commanded_position_nla}
    x_c[k]-x_c[k-1] =\Delta x_r[k](\alpha[k]-\alpha[k-1])  
\end{equation} 
The convergence rate of the sigmoid function in Equation~\eqref{eqn:nonlinear_accommodation_force_feedback} can be controlled by the accommodation term $K_a$. Using the asymptotic convergence of $\alpha[k]$, we have that $\exists$ a $\mathcal{K}\in \mathbb{Z}$ such $\forall $ $k>\mathcal{K}$, $||\alpha[k]-\alpha[k-1]||_2\leq \hat{\epsilon}$. Then we can use this to re-write Equation~\eqref{eqn:commanded_position_nla} as the following:
\begin{align}
    ||x_c[k]-x_c[k-1]||_2 & \leq\Delta x_r||\alpha[k]-\alpha[k-1]||_2 \nonumber \\
        &\leq \tilde{\epsilon}\label{eqn:x_conv_nla}
\end{align}

where, $\tilde{\epsilon}=\Delta x_r\hat{\epsilon} $. Convergence of $x_c$ follows from the fact that the $\tilde{\epsilon}$ can be made arbitrarily small. Then using Assumption~\ref{assumption1} and Equation~\eqref{eqn:x_conv_nla}, we can show that   $||f[k]-f[k-1]||_2\leq \epsilon$.
 
\end{proof}
\end{thm}

The assumption regarding the existence of $\gamma$ and $K_a$ is not very strict. In practice, we found that we were able to find an interval for $\gamma \in (0.3,0.6)$ for which our infinite sum converged. The plots shown in Figure~\ref{fig:force_convergence}a were obtained with $\gamma =0.35$. The above provides us a solution to the first problem that was presented in Section~\ref{sec:problem_statement}. In the next section, we analyze the data collected using the proposed force controller and present the design of a learning-based controller for peg insertion.

\section{Learning Predictive Model for Misalignment}\label{sec:predictive_model}
In this section, we analyze contact wrench data to understand the dependence of the contact wrench on the misalignment. To provide a complete understanding of this relationship, we analyze the force signature data that is collected during an initial training phase. The purpose of this model is to predict misalignment based on the force signature which is characteristic of a certain contact configuration. 

To learn the predictive model, we collect training data consisting of the force signature for the contact configuration by a known amount of misalignment. We use the accommodation controller that we presented earlier during data collection to ensure safe interaction during this exploration phase. Furthermore, we also ensure that we measure the force signature for a given misalignment at a quasi-steady state, when it has converged to an asymptotic value, which simplifies the learning problem.

\subsection{Data Collection}\label{subsec:data_collection}
To learn a predictive model for correcting misalignment, we collect data by introducing misalignment in the position of the peg with respect to the hole. The work in this paper only considers planar misalignment between the peg and the hole. Consequently, we introduce misalignment in the $x$ and $y$ axes from the known hole location. The misalignment is sampled from a uniform distribution from the interval $[-3, 3]$ mm. This interval was chosen, because the deep learning-based hole detection method we used is able to achieve similar accuracy in the estimated position of the hole. With the added misalignment in the position of the peg, any insertion attempt leads to a contact formation between the peg and the hole environment. The contact formation leads to a force signature that is observed through the F/T sensor mounted at the wrist of the robot (see Figure~\ref{fig:exp_setup}). For every episode of data collection, the robot follows the insertion trajectory, and records the force measurements measured through the F/T sensor for the resulting contact formation. Thus, we collect a data set where we store the misalignment as well as the measured force signature corresponding to the misalignment. We use a Mitsubishi Electric Factory Automation (MELFA) RV-$5$AS-D
 Assista $6$-DoF  arm (see Figure~\ref{fig:exp_setup}) for the experiments. The robot has pose repeatability of $\pm 0.03$mm. The robot is equipped with Mitsubishi Electric F/T sensor $1$F-FS$001$-W$200$ (see Figure~\ref{fig:exp_setup}). In the initial set of experiments, we also verify that Assumption~\ref{assumption1} is valid for our robotic setup. 

\subsection{Numerical Analysis for Convergence}
\begin{figure}
    \centering
    \includegraphics[width=0.75\columnwidth]{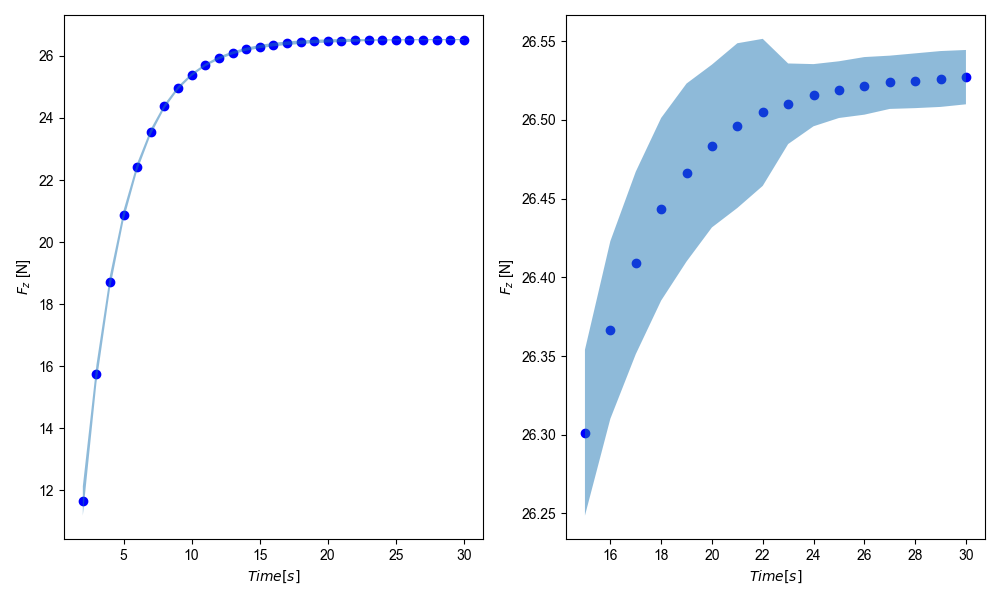} %
    \caption{Average values of the vertical forces computed every consecutive $1[sec]$ on the trajectory. Left: shows the mean and the $95\%$ confidence interval of these values for all the experiments. Right: zooms in only the last $15[sec]$ of the trajectory.}
    \label{fig:mean_analysis_lac}
\end{figure}

\begin{figure}
    \centering
    \includegraphics[width=0.75\columnwidth]{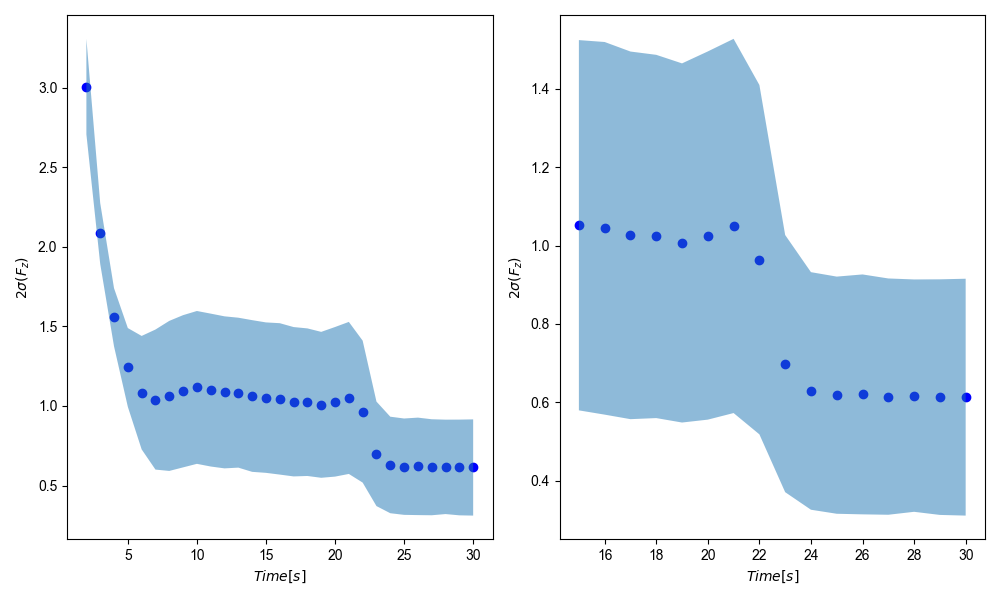} %
    \caption{Standard deviation values of the vertical forces computed every consecutive $1[sec]$ on the trajectory. Left: shows the mean and the $95\%$ confidence interval of these values for all the experiments. Right: zooms in only the last $15[sec]$ of the trajectory.}
    \label{fig:std_analysis_lac}
\end{figure}

\begin{figure}
    \centering
    \includegraphics[width=0.75\columnwidth]{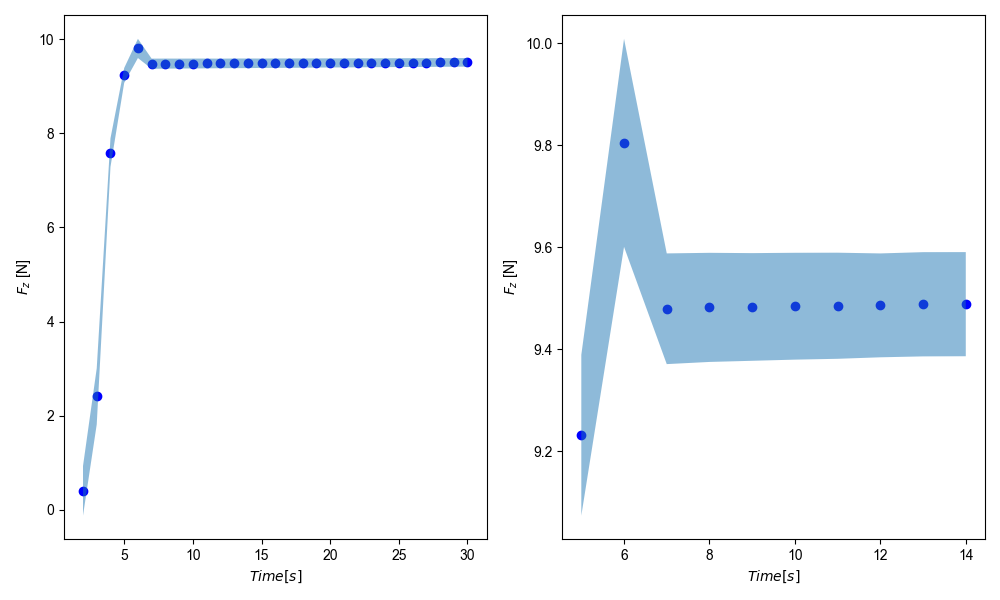} %
    \caption{Average values of the vertical forces computed every consecutive $1[sec]$ on the trajectory. Left: shows the mean and the $95\%$ confidence interval of these values for all the experiments. Right: zooms in only the first $15[sec]$ of the trajectory.}
    \label{fig:mean_analysis_nlac}
\end{figure}

\begin{figure}
    \centering
    \includegraphics[width=0.75\columnwidth]{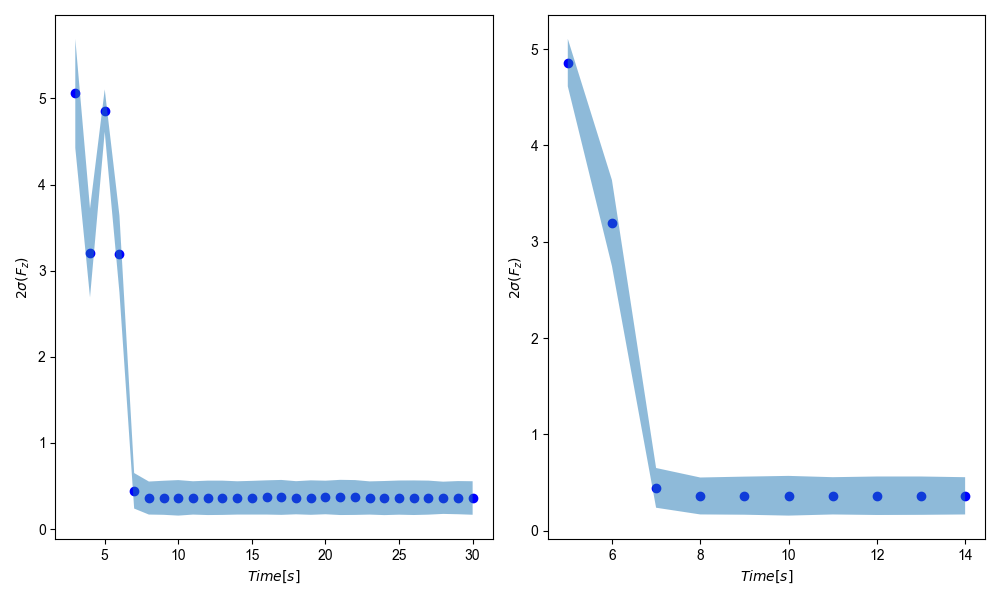} %
    \caption{Standard deviation values of the vertical forces computed every consecutive $1[sec]$ on the trajectory. Left: shows the mean and the $95\%$ confidence interval of these values for all the experiments. Right: zooms in only the first $15[sec]$ of the trajectory.}
    \label{fig:std_analysis_nlac}
\end{figure}

We analyse the convergence properties of the proposed controllers. In Figures~\ref{fig:mean_analysis_lac} and~\ref{fig:std_analysis_lac}, the statistics of the force signature measured by the F/T sensor along the vertical direction have been reported for regular time intervals on all $1,200$ experiments described in Section~\ref{subsec:data_collection} for the linear controller. Similarly, we report these quantities for the nonlinear controller in Figures~\ref{fig:mean_analysis_nlac} and~\ref{fig:std_analysis_nlac}. 

In particular, we have computed the mean, $\Bar{F}_z$, and twice the standard deviation, $2\sigma( F_z )$, for each time interval of $1 [sec]$ along the trajectory and for all the 1200 experiments. The mean and the $95\%$ confidence interval of these two statistics are reported in Figures~\ref{fig:mean_analysis_lac} and~\ref{fig:std_analysis_lac}, respectively for the linear and in Figures~\ref{fig:mean_analysis_nlac} and~\ref{fig:std_analysis_nlac}, for the nonlinear controller, respectively.

The practical purpose of this analysis is to be able to decide online when the controller has converged to a stable value of the vertical forces as soon as possible. The criterion we selected to decide the convergence of the system is based on the changes we can observe in the 4 statistics we have described above: the mean, $\Bar{F}_z$, the standard deviation, $2\sigma( F_z )$, the mean of the standard deviation, $2\mathbf{E}[\sigma( F_z )]$, and the standard deviation of the standard deviation $2\sigma( 2\sigma( F_z ) )$. We then took the difference of each of these statistics with respect to time intervals, $\Delta X^i$ with $i = \{1,\ldots,4\}$ where $X^i$ is one of the four statistics and $\Delta X^i_k =X^i_k - X^i_{k-1} $. We declare that the system converged if 
\begin{equation}\label{eq:criterion_conv}
    \wedge_{i=1}^4 \Delta X^i_k < \eta_{\text{th}},
\end{equation}
for 2 consecutive time intervals $k$, where $\wedge$ is in the end operator. Basically we are requesting for the time interval where the changes for all the statics are less than a predetermined threshold $\eta_{\text{th}}$.

Criterion \eqref{eq:criterion_conv} applied to the linear controller, see signals in Figures~\ref{fig:mean_analysis_lac},~\ref{fig:std_analysis_lac}, output that the controller converged after $25s$. Analogously, for the nonlinear controller, see signals in Figures~\ref{fig:mean_analysis_nlac},~\ref{fig:std_analysis_nlac}, the controller converged after $9s$.
Note that the confidence intervals never go to zero because of the measurement noise. This empirical analysis confirms the theoretical convergence results shown in Section~\ref{sec:controller_design}. Therefore, the classifiers can be computed based on values at convergence without having to wait for the end of the experiment.

\subsection{Model Learning Performance}\label{subsec:ML_models}
We train classification and regression models using the collected contact force data to learn predictive models for direction and magnitude of misalignment. We use the results from the previous section to decide on the convergence and use the convergence criterion to decide when to stop collecting data during the contact formation. We use $\eta_{\text{th}}$ found in the last section for collecting the force signature. We then train a classification and a regression model to learn the direction and the magnitude of error respectively, to understand the efficacy of the models. The results of classification and regression are shown in Tables~\ref{table:GPC_prediction} and~\ref{table:GPR_prediction} (see the results with full features). Note that we are able to achieve better result with the linear accommodation controller; however, the nonlinear controller can predict these faster than the linear controller. Another point to notice is that we are able to achieve good RMSE scores for both controllers -- the linear controller is better than the nonlinear. However, this problem requires that we should be able to predict the directions accurately. The regression model with the linear accommodation controller is able to predict the direction with accuracy $98\%$ and $94\%$ in the $x$ and $y$ axis respectively. With the nonlinear controller, we achieve an accuracy of $97\%$ and $92\%$. Notice that overall the linear controller is able to achieve higher accuracy. This might be due to higher interaction forces which leads to less noise in the force signatures. This might be one of the reasons for the slightly better performance of the linear controller. 


\begin{table}[t]
    \caption{{Classification accuracy in prediction of direction of misalignment along X and Y axis using Gaussian process classifiers.}}
    \centering
    \begin{tabular}{c|c|c|c|c}
     \multirow{3}{*}{Axis} & \multicolumn{4}{c}{Classification Accuracy (higher is better)} \\
     \cline{2-5}
     & \multicolumn{2}{c}{Linear Controller} & \multicolumn{2}{c}{NonLinear Controller} \\
     \cline{2-5}
     & Full Feat & Reduced Feat & Full Feat & Reduced Feat\\
         \hline\hline  X & 0.9964&0.9916 &0.9928 &0.9916\\
         \hline Y & 0.939 &0.949 & 0.92&0.920\\
    \end{tabular}
    \label{table:GPC_prediction}
\end{table}


\begin{table}[t]
    \caption{{Regression accuracy in prediction of magnitude of misalignment along X and Y axis using Gaussian process regression.}}
    \centering
    \begin{tabular}{c|c|c|c|c}
     \multirow{3}{*}{Axis} & \multicolumn{4}{c}{RMSE [mm] (lower is better)} \\
     \cline{2-5}
     & \multicolumn{2}{c}{Linear Controller} & \multicolumn{2}{c}{NonLinear Controller} \\
     \cline{2-5}
     & Full Feat & Reduced Feat & Full Feat & Reduced Feat\\
         \hline\hline  X & 0.59 &0.61 &0.59 &0.55\\
         \hline Y & 0.83 &0.67 & 0.92 &0.72\\
    \end{tabular}
    \label{table:GPR_prediction}
\end{table}

\subsection{Feature Importance}
\begin{figure}
    \centering
    \includegraphics[width=0.40\textwidth]{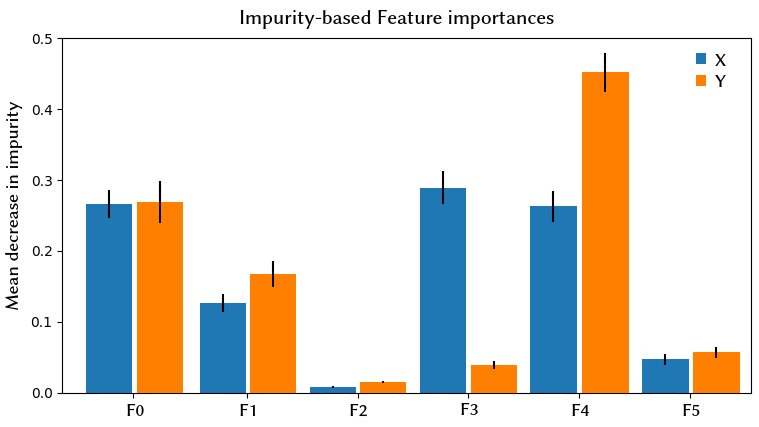} %
    \caption{Feature importance provided by the fitted attribute computed as the mean and standard error of accumulation of the impurity decrease within each tree with a forest of trees. Blue and orange bars describe which features are relevant for learning the predictive model for X and Y direction  misalignment, respectively.}
    \label{fig:feature_analysis}
\end{figure}

We use feature importance analysis to describe which features are relevant for learning the predictive model for misalignment from force observations. Feature analysis can help with a better understanding of the problem. In particular, we use a forest of trees to evaluate the importance of the force features on the classification task~\cite{liaw2002classification}. We consider the Cartesian force signals and the corresponding moment signals from the F/T sensor to obtain the wrench signal $[f_x,f_y,f_z, m_x, m_y, m_z]$, which we use as features for identifying the hole misalignment. The fitted attribute provides feature importance, and they are computed as the mean and standard error of accumulation of the impurity decrease within each tree. We observe that $x$ and $y$ Cartesian force signals $(F_0, F_1)$ and moments $(F_3, F_4)$ are found important for the classification task, where $z$ Cartesian force signals and moments $(F_2, F_5)$ are unimportant. In figure~\ref{fig:feature_analysis}, the bars are the feature importance of the forest, along with their inter-trees variability represented by the error bars. This agrees with the physical intuition about the insertion-- since the forces in the $z$ direction are constant for all trials, it should not be helpful in providing any discriminating information for class separation. Similarly, the contact formation during insertion attempts should not lead to any moment in $z$, and thus this information is also not useful for making misalignment decisions. We repeat the classification and regression modeling with the reduced feature sets and the results are listed in Tables~\ref{table:GPC_prediction} and~\ref{table:GPR_prediction} (see the reduced feature results). It can be observed that we can achieve better performance than using the full force signature for learning a predictive model. This shows the effectiveness of feature selection and that we are able to do better than using the entire $6$-dimensional wrench vector.

\section{Results for Insertion}\label{sec:results}
In this section, we present design of an insertion policy using the predictive model using the force signature data that was described in the previous section. For completeness of presentation, we present brief details of the deep learning-based hole detection framework which is used for testing the performance of the force controller proposed in the paper. This vision module is used in the proposed work to perform hole-detection-based insertion. The approach is based on our previous work presented in~\cite{jain2022vision}. In this section, we first present details of the vision module that we use for hole detection, and then present results for insertion using the learned predictive models.

\subsection{Vision System for Hole Detection}

\begin{figure}
    \centering
    \includegraphics[width=0.5\textwidth]{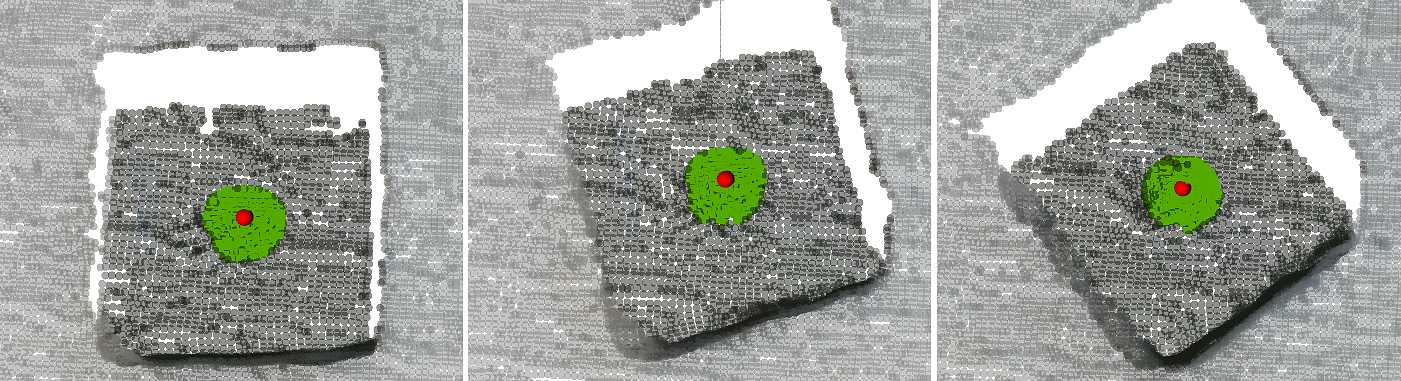} %
    \caption{Samples of the visual hole detection shown on the point cloud of the object. Detected hole locations are indicated in red and the masks are shown in green.}
    \label{fig:hole_detection}
\end{figure}

We choose a supervised learning approach to detect the hole location from visual sensory data obtained from an RGB-D sensor (Intel Realsense, D435) camera. Using traditional computer vision approaches to detect hole location with unknown object pose might lead to false positives (e.g., template matching~\cite{briechle2001template} or the Hough circle transform~\cite{yuen1990comparative}). We implement the Mask R-CNN~\cite{he2017mask} deep learning architecture for instance-level segmentation to detect hole locations. Our classification setup has two classes, one for the background and one for the hole location. The network prediction identifies the resulting segmentation masks for hole locations. We performed transfer learning from the MS COCO dataset pre-trained weights in a supervised manner. For the learning dataset, we captured 300 images of size 640×480 at different distances. We annotated the data to indicate hole pixels with the labelme~\cite{russell2008labelme} annotation tool. At inference time, we utilize the detected segmentation mask of the hole location to compute the corresponding registered point cloud data points. The output from the approach is the estimate of the 3D hole location from the visual sensory data. Figure~\ref{fig:hole_detection} shows some qualitative samples of hole detection approach on point cloud of the test object. 

\subsection{Insertion Using Force Feedback Models}\label{subsec:insertion_results}
In this section, we present results from performing insertion with the trained force feedback models in the presence of error in the detection of hole location. We use the vision module to detect the approximate location of hole in the environment of the robot. Compared to our previous work proposed in~\cite{jain2022vision, jha2021imitation}, we experiment with parts with tighter tolerances to test the performance of the force controller. In particular, we use parts with tolerance of approximately $1$ mm. To test the performance of the integrated force with the vision-based hole detection, we move the object in the field of view of the RGB-D sensor (see Figure~\ref{fig:exp_setup}). The robot is now asked to perform insertion based on the estimate of the DL method for hole location. We find that the DL-based method is fairly accurate for reaching in the vicinity of the hole location. Then we use the learned force controller for performing insertion, overcoming any misalignment. We use the classification prediction by the trained classifiers to move by a unit step of $0.5$ mm in the predicted direction while maintaining contact with the object surface. This is repeated till either the robot succeeds in insertion or diverges more than $5$ mm. The robot is given a maximum of $10$ correction attempts. We measure the number of corrections made by the linear and non-linear controllers. We move the object to $20$ random location in the view of the camera and attempt insertion. We observe that the ML model with linear controller is able to achieve $100\%$ success rate with average number of corrections to be $1.2$ while the ML model with non-linear controller achieves $95\%$ success rate (1 failure case out of 20 attempts) with an average correction rate of $1.8$ per successful attempt. A more thorough analysis of the controller performance is left to an extended version of the paper. 


\section{Conclusions and Future Work}\label{sec:conclusions}
In this paper, we presented the design and analysis of accommodation controllers for contact interaction during assembly operations and their use in adaptive assembly strategies based on machine learning. Most assembly operations with tight tolerances result in complex contact formations which might damage the parts being assembled. Ensuring safe operation of robots requires the design of force feedback controllers that can ensure limited contact forces in the presence of sustained contacts. In this paper, we presented two designs of generalized accommodation controllers that use force feedback during contact interaction to ensure limited contact forces. We presented analysis of these controllers to show convergence of contact forces under the assumption of constant velocity of the underlying reference trajectory. We presented results from different machine learning models which were trained using different signal statistics, and compared them to find an optimal signal feature. Finally, we used the trained model to perform insertion using a DL-based vision algorithm for hole detection. We show that we are able to achieve $100\%$ success rate for insertion using the proposed controllers and using the DL-based vision system for detecting hole location with tolerances tighter than $1$ mm.

In the future, we will perform more rigorous comparison between the linear and nonlinear controllers at different operating velocities of the robot, and find the best operating conditions which leads to fastest insertion times and highest success rate. 


\bibliographystyle{IEEEtran}
\bibliography{references}

\end{document}